\def\year{2019}\relax
\documentclass[letterpaper]{article} %DO NOT CHANGE THIS
\usepackage{aaai19}  %Required
\usepackage{times}  %Required
\usepackage{helvet}  %Required
\usepackage{courier}  %Required
\usepackage{url}  %Required
\usepackage{hyperref}

\usepackage{graphicx}  %Required
\usepackage{amsmath}
\usepackage{color}
\usepackage{float}
\usepackage{epstopdf}

\newcommand{\citet}[1]
{\citeauthor{#1} ̃\shortcite{#1}}
\newcommand{\citep}{\cite}

\usepackage{macros}
\usepackage{amsfonts}       % blackboard math symbols
\usepackage{nicefrac}       % compact symbols for 1/2, etc.
\usepackage{microtype}   
\usepackage{graphicx}
\usepackage{capt-of}

\begin{document}
% The file aaai.sty is the style file for AAAI Press 
% proceedings, working notes, and technical reports.
%
\title{Towards Non-saturating Recurrent Units for Modelling Long-term Dependencies}
\author{\qquad Sarath Chandar$^{1,2}$\thanks{Equal Contribution}\thanks{Corresponding author: apsarathchandar@gmail.com.}\qquad~~ Chinnadhurai Sankar$^{1}$\footnotemark[1]\qquad\qquad Eugene Vorontsov$^{1}$\thanks{Work done while at Microsoft Research.} \\\\
\textbf{\Large{Samira Ebrahimi Kahou}}$^{3}$ \qquad\qquad \textbf{\Large{Yoshua Bengio}\thanks{CIFAR Fellow.}}$^{1}$\\\\
    $^1$\Large{Mila, Universit\'e de Montr\'eal} \qquad
    $^2$\Large{Google Brain}\qquad
    $^3$\Large{Microsoft Research}
}
\maketitle
\begin{abstract}
Modelling long-term dependencies is a challenge for recurrent neural networks. This is primarily due to the fact that gradients vanish during training, as the sequence length increases. Gradients can be attenuated by transition operators and are attenuated or dropped by activation functions. Canonical architectures like LSTM alleviate this issue by skipping information through a memory mechanism. We propose a new recurrent architecture (Non-saturating Recurrent Unit; NRU) that relies on a memory mechanism but forgoes both saturating activation functions and saturating gates, in order to further alleviate vanishing gradients. In a series of synthetic and real world tasks, we demonstrate that the proposed model is the only model that performs among the top 2 models across all tasks with and without long-term dependencies, when compared against a range of other architectures.
\end{abstract}

\section{Introduction}

Vanishing and exploding gradients remain a core challenge in the training of recurrent neural networks. While exploding gradients can be alleviated by gradient clipping or norm penalties~\citep{pascanu2013}, vanishing gradients are difficult to tackle. Vanishing gradients prevent the network from learning long-term dependencies \cite{hochreiter1991,bengio1994} and arise when the network dynamics have attractors necessary to store bits of information over the long term \citep{bengio1994}. At each step of backpropagation, gradients in a recurrent neural network can grow or shrink across linear transition operators and shrink across nonlinear activation functions.

Successful architectures, like the LSTM \cite{hochreiter1997} and GRU \cite{cho2014}, alleviate vanishing gradients by allowing information to skip transition operators (and their activation functions) via an additive path that serves as memory. Both gates and transition operators use bounded activation functions (sigmoid, tanh). These help keep representations stable but attenuate gradients in two ways: they are contractive and they saturate at the bounds. Activation functions whose gradients contract as they saturate are typically referred to as saturating nonlinearities; in that sense, rectified linear units (ReLU) \cite{nair2010rectified} are non-saturating and thus help reduce vanishing gradients in deep feed-forward networks \cite{krizhevsky2012,xu2015empirical}.

Saturating activation functions still dominate in the literature for recurrent neural networks, in part because of the success of LSTM on sequential problems \citep{sutskever2014,vinyals2015,merity2017,mnih2016}. As these functions saturate, gradients contract. In addition, while saturated gates may push more information through memory, improving gradient flow across a long sequence, the gradient to the gating units themselves vanishes. We expect that non-saturating activation functions and gates could improve the modelling of long-term dependencies by avoiding these issues. We propose a non-saturating recurrent unit (NRU) that forgoes both saturating activation functions and saturating gates. We present a new architecture with the rectified linear unit (ReLU)  and demonstrate that it is well equipped to process data with long distance dependencies, without sacrificing performance on other tasks.

\section{Background}
\subsection{Vanishing Gradient Problem in RNNs}
Recurrent Neural Networks are the most successful class of architectures for solving sequential problems. A vanilla RNN, at any time step $t$, takes $\vx_t$ as input and updates its hidden state as follows:
\begin{align}
    \vz_t &= \mW\vh_{t-1} + \mU\vx_t\\
    \vh_t &= f(\vz_t) ,\\
\end{align}
where $\mW$ and $\mU$ are the recurrent and the input weights of the RNN respectively and $f(.)$ is any non-linear activation function like a sigmoid or tanh. Consider a sequence of length $T$ with loss $\LL$ computed at the end of the sequence. To compute $\frac{\partial\LL}{\partial{W_{ij}}}$ at time step $t$, we should first compute $\frac{\partial\LL}{\partial\vh_t}$ using the following chain rule:
\begin{align}
    \frac{\partial\LL}{\partial\vh_t} &= \frac{\partial\LL}{\partial\vh_T} \frac{\partial\vh_T}{\partial\vh_t}\\
    &= \frac{\partial\LL}{\partial\vh_T} \prod_{k=t}^{T-1} \frac{\partial\vh_{k+1}}{\partial\vh_k}\\
    &= \frac{\partial\LL}{\partial\vh_T} \prod_{k=t}^{T-1} \left(\text{diag}[f'(\vz_k)] \mW \right) . \label{repW}
\end{align}
For a very long sequence length $T$, repeated multiplication of $\mW$ in Equation \ref{repW} can cause exponential growth or decay of backpropagated gradients. Exploding gradients are caused by eigenvalues greater than one in $\mW$. Conversely, eigenvalues less than one lead to vanishing gradients. The second cause of vanishing gradients is the activation function $f$. If the activation function is a saturating function like a sigmoid or tanh, then its Jacobian $\text{diag}[f'(\vz_k)]$ has eigenvalues less than or equal to one, causing the gradient signal to decay during backpropagation. Long term information may decay when the spectral norm of $\text{diag}[f'(\vz_k)]$ is less than 1 but this condition is actually necessary in order to store this information reliably \citep{bengio1994,pascanu2013}. While gradient clipping can limit exploding gradients, vanishing gradients are harder to prevent and so limit the network's ability to learn long term dependencies \citep{hochreiter1991,bengio1994}.

\subsection{Existing Solutions}

A key development in addressing the vanishing gradient issue in RNNs was the introduction of gated memory in the Long Short Term Memory (LSTM) network \cite{hochreiter1997}. The LSTM maintains a cell state that is updated by addition rather than multiplication and serves as an ``error carousel" that allows gradients to skip over many transition operations. 

There have been several improvements proposed in the literature to make LSTM learn longer term dependencies. \citet{Gers2000} proposed to initialize the bias of the forget gate to 1 which helped in modelling medium term dependencies when compared with zero initialization. This was also recently highlighted in the extensive exploration study by \citet{jozefowicz2015}. \citet{tallec2018} proposed chrono-initialization to initialize the bias of the forget gate ($\vb_f$) and input gate ($\vb_i$). Chrono-initialization requires the knowledge of maximum dependency length ($T_{\text{max}}$) and it initializes the gates as follows:
\begin{align}
    \vb_f &\sim \text{log}(\UU[1, T_{\text{max}} - 1])\\
    \vb_i &= -\vb_f .
\end{align}
This initialization method encourages the network to remember information for approximately $T_{\text{max}}$ time steps. While these forget gate bias initialization techniques encourage the model to retain information longer, the model is free to unlearn this behaviour.

The Gated Recurrent Unit (GRU) \citep{cho2014} is a simplified version of the LSTM (with fewer gates) which works equally well \cite{Chung2014}. Recently, \citet{vander2018} proposed a forget-gate only version of LSTM called JANET. JANET outperforms the LSTM on some long-term dependency tasks. We suspect that this may be because JANET omits a few saturating gates in the LSTM which are not crucial for the performance of the model. In that sense, JANET is similar to the proposed NRU. However, the NRU uses non-saturating gates and hence should have better gradient flow than JANET.

The non-saturating ReLU activation function should reduce the incidence of vanishing gradients in a vanilla RNN but can make training unstable through unbounded growth of the hidden state. Better initialization of the recurrent weight matrix has been explored in the literature for such RNNs. \citet{le2015} proposed initializing the recurrent weight matrix using the identity matrix. Orthogonal initialization of recurrent weight matrices has been studied by \citet{henaff2016}. These initialization techniques, like bias initialization in the LSTM, encourage better gradient flow at the beginning of the training. \citet{vorontsov2017} proposed various ways to parameterize the recurrent weight matrix so that it could stay orthogonal or nearly orthogonal. They observed that allowing slight deviation from orthogonality helps improve model performance and convergence speed. 

The idea of using unitary matrices for recurrent weight matrices has been explored in \cite{arjovsky2016,wisdom2016,jing2017}. While \cite{arjovsky2016,jing2017} parameterize the matrix to be unitary, \cite{wisdom2016} reprojects the matrix after every gradient update. Orthogonal and unitary matrices do not filter out information, preserving gradient norm but making forgetting depend on a separate mechanism such as forget gates. This has been explored with saturating gates in \cite{jing2017b,dangovski2017}.

A non-gated Statistical Recurrent Unit (SRU) was proposed in \cite{oliva2017}. SRU uses only a ReLU activation function and recurrent multi-scale summary statistics. The summary statistics are calculated using exponential moving averages which might shrink the gradients. We compare NRU with SRU in our experiments.

Other related lines of work which aim to model long-term dependencies in RNNs are Memory Augmented Neural Networks (MANNs) \cite{graves2014,graves2016,gulcehre2018,gulcehre2017}. \cite{graves2014,graves2016} have continuous addressing schemes for the external memory, which also suffer from the vanishing gradient problem. \cite{gulcehre2018,gulcehre2017} attempts to address this issue by moving to a discrete addressing mechanism so that gradients could flow better. However, they use approximate gradient methods like REINFORCE \cite{williams92} or the Gumbel-softmax trick \cite{maddison2016,jang2016} which are either difficult to train or to scale. All these architectures use saturating activation functions and saturating gates for memory control.

Data flow normalization techniques like layer normalization \cite{ba2016} and recurrent batch normalization \cite{cooijmans2016} were also proposed in the literature to improve the gradient flow in recurrent architectures. \citet{noisy2016} proposed injecting noise in the saturating gates to tend to recover gradient flow at saturation.

\section{Non-saturating Recurrent Unit (NRU) Network}

Saturating gating functions introduce a trade-off where distant gradient propagation may occur at the cost of vanishing updates to the gating mechanism. We seek to avoid this trade-off between distant gradient propagation but not updateable gates (saturated gates) and short gradient propagation but updateable gates (non-saturated gates). To that end, we propose the Non-saturating Recurrent Unit --- a gated recurrent unit with no saturating activation functions.

\subsection{Model}

At any time step $t$, NRU takes some input $\vx_t$ and updates its hidden state as follows:
\begin{align}
\vh_t &= f(\mW_h\vh_{t-1} + \mW_i\vx_t + \mW_c\vm_{t-1})
\end{align}
where $\vh_{t-1} \in \mathbb{R}^h$ is the previous hidden state, $\vm_{t-1} \in \mathbb{R}^m$ is the previous memory cell, and $f$ is a ReLU non-linearity. The memory cell in NRU is a flat vector similar to the cell vector in LSTM. However, in NRU, the hidden state and the memory cell need not be of the same size. We allow the memory cell to be larger than the hidden state to hold more information. 

At every time step $t$, memory cell is updated as as follows:
\begin{align}
\vm_t = \vm_{t-1} + \sum_{k_1} \alpha_i \vv^{w}_{i} - \sum_{k_2} \beta_i \vv^{e}_{i}
\label{eq:memory_update_eq}
\end{align}
where
\begin{itemize}
    \item $\vm_t \in \mathbb{R}^m$ is the memory vector at time $t$.
    \item $k_1$ is the number of writing heads and $k_2$ is the number of erasing heads. We often set $k_1 = k_2 = k$.
    \item $\vv^{w}_i \in \mathbb{R}^m$ is a normalized vector which defines where to write in the memory. Similarly $\vv^{e}_i \in \mathbb{R}^m$ is a normalized vector which defines where to erase in the memory.
    \item $\alpha_i$ is a scalar value which represents the content which is written in the memory along the direction of $\vv^w_i$. Similarly, $\beta_i$ is a scalar value which represents the content which is removed from the memory along the direction of $\vv^e_i$.
\end{itemize}

Intuitively, the network first computes a low-dimensional projection of the current hidden state by generating a $k$-dimensional $\alpha$ vector. Then it writes each unit of this $k$-dimensional vector along $k$ different basis directions ($\vv^w_i$s). These $k$ basis directions specify \textit{where to write} as continuous-valued vectors and hence there is no need for saturating non-linearities or discrete addressing schemes. The network also similarly erases some content from the memory (by using $\beta$ and $\vv^e_{i}$s).

Scalars $\alpha_i$ and $\beta_i$ are computed as follows:
\begin{align}
    \alpha_i &= f_{\alpha}(\vx_t, \vh_t, \vm_{t-1})\\
    \beta_i &= f_{\beta}(\vx_t, \vh_t, \vm_{t-1})
\end{align}
where $f_{\alpha}$ and $f_{\beta}$ are linear functions followed by an optional ReLU non-linearity. Vectors $\vv^w_i$ and $\vv^e_i$ are computed as follows:
\begin{align}
    \vv^w_i &= f_{w}(\vx_t, \vh_t, \vm_{t-1})\\
    \vv^e_i &= f_{e}(\vx_t, \vh_t, \vm_{t-1})
\end{align}
where $f_w$ and $f_e$ are linear functions followed by an optional ReLU non-linearity followed by an explicit normalization. We used $L_{5}$ normalization in all of our experiments.

Each $\vv$ vector is $m$-dimensional and there are $2k$ such vectors that needs to be generated. This requires a lot of parameters which could create problems both in terms of computational cost and in terms of overfitting. We reduce this large number of parameters by using the following outer product trick to factorize the $\vv$ vectors:
\begin{align}
    \vp^w_i &= f_{w_1}(\vx_t, \vh_t, \vm_{t-1}) \in \mathbb{R}^{\sqrt{m}}\\
    \vq^w_i &= f_{w_2}(\vx_t, \vh_t, \vm_{t-1}) \in \mathbb{R}^{\sqrt{m}}\\
    \vv^w_i &= g(\text{vec}(\vp^w_i {\vq^w_i}^T))
\end{align}
where $f_{w_1}$ and $f_{w_2}$ are linear functions, vec() vectorizes a given matrix, $g$ is an optional ReLU non-linearity followed by an explicit normalization. Thus, instead of producing an $m$-dimensional vector for the direction, we only need to produce a $2\sqrt{m}$-dimensional vector which requires substantially fewer parameters and scales more reasonably. We can further reduce the number of parameters by generating $2\sqrt{km}$-dimensional vector instead of $2k\sqrt{m}$-dimensional vector and then use the outer product trick to generate the required $km$-dimensional vector. We do this trick in our implementation of NRU.

If there is no final ReLU activation function while computing $\alpha$, $\beta$, and $\vv$, then there is no distinction between write and erase heads. For example, either $\beta_i$ or $\vv^e_i$ could become negative, changing the erasing operation into one that adds to the memory. Having an explicit ReLU activation in all these terms forces the architecture to use writing heads to add information to the memory and erasing heads to erase information from the memory. However, we treat this enforcement as optional and in some experiments, we let the network decide how to use the heads by removing these ReLU activations.

\subsection{Discussion}

In vanilla RNNs, the hidden state acts as the memory of the network. So there is always a contradiction between stability (which requires small enough $\mW_h$ to avoid explosions) and long-term storage (which requires high enough $\mW_h$ to avoid vanishing gradients). 
This contradiction is exposed in a ReLU RNN which avoids the saturation from gates at the cost of network stability. LSTM and GRU reduce this problem by introducing memory in the form of information skipped forward across transition operators, with gates determining which information is skipped. However, the gradient on the gating units themselves vanishes when the unit activations saturate. Since distant gradient propagation across memory is aided by gates that are locked to ON (or OFF), this introduces an unfortunate trade-off where either the gate mechanism receives updates or gradients are skipped across many transition operators.

NRU also maintains a separate memory vector like LSTM and GRU. Unlike LSTM and GRU, NRU does not have saturating gates. Even if the paths purely through hidden states $h$ vanish (with small enough $\mW_h$), the actual long term memories are stored in memory vector $\vm$ which has additive updates and hence no vanishing gradient.

Although the explicit normalization in vectors $\vv^w_i$ and $\vv^e_i$  may cause gradient saturation for large values, we focus on the nonlinearities because of their outsized effect on vanishing gradients. In comparison, the effect of normalization is less pronounced than that of explicitly saturating nonlinearities like tanh or sigmoid which are also contractive based on our experiments. Also, normalization has been employed to avoid the saturation regimes of these nonlinearities \citep{batchnorm,cooijmans2016}. Furthermore, we observe that forgoing these nonlinearities allows the NRU to converge much faster than other models.

\section{Experiments}
In this section, we compare the performance of NRU networks with several other RNN architectures in three synthetic tasks (5 including variants) and two real tasks. Specifically, we consider the following extensive list of recurrent architectures: RNN with orthogonal initialization (RNN-orth), RNN with identity initialization (RNN-id), LSTM, LSTM with chrono initialization (LSTM-chrono), GRU, JANET, SRU, EURNN \cite{jing2017}, GORU \cite{jing2017b}. We used the FFT-like algorithm for the orthognal transition operators in EURNN and GORU as it tends to be much faster than the tunable algorithm, both proposed by \cite{jing2017}. While these two algorithms have a low theoretical computational complexity, they are difficult to parallelize.

We mention common design choices for all the experiments here: RNN-orth and RNN-id were highly unstable in all our experiments and we found that adding layer normalization helps. Hence all our RNN-orth and RNN-id experiments use layer normalization. We did not see any significant benefit in using layer normalization for other architectures and hence it is turned off by default for other architectures. NRU with linear writing and erasing heads performed better than with ReLU based heads in all our experiments except the language model experiment. Hence we used linear writing and erasing heads unless otherwise mentioned. We used the Adam optimizer \cite{kingma2014} with a default learning rate of 0.001 in all our experiments. We clipped the gradients by norm value of 1 for all models except GORU and EURNN since their transition operators do not expand norm. We used a batch size of 10 for most tasks, unless otherwise stated.

Table-\ref{toppers} summarizes the results of this section. Out of 10 different architectures that we considered, NRU is the only model that performs among the top 2 models across all 7 tasks. The code for NRU Cell is available at \url{https://github.com/apsarath/NRU}. 

\begin{table}[htbp]
\centering
\caption{Number of tasks where the models are in top-1 and top-2. Maximum of 7 tasks. Note that there are ties between models for some tasks so the column for top-1 performance would not sum up to 7.}
\begin{tabular}{l|l|l}
\hline
\textbf{Model} & \textbf{in Top-1} & \textbf{in Top-2}\\
\hline
\hline
RNN-orth & 1 & 1  \\
RNN-id & 1 & 1  \\
LSTM & 1 & 1  \\
LSTM-chrono & 1  & 1  \\
GRU & 1 & 2  \\
JANET & 1 & 3   \\
SRU & 1 & 1  \\
EURNN & 2 & 3  \\
GORU & 0 & 1  \\
NRU & \textbf{4} & \textbf{7}\\
\hline
\end{tabular}
\label{toppers}
\end{table}

\subsection{Copying Memory Task}

\begin{figure*}[h]
\centering
\begin{minipage}{.5\textwidth}
\centering
\includegraphics[scale=0.55]{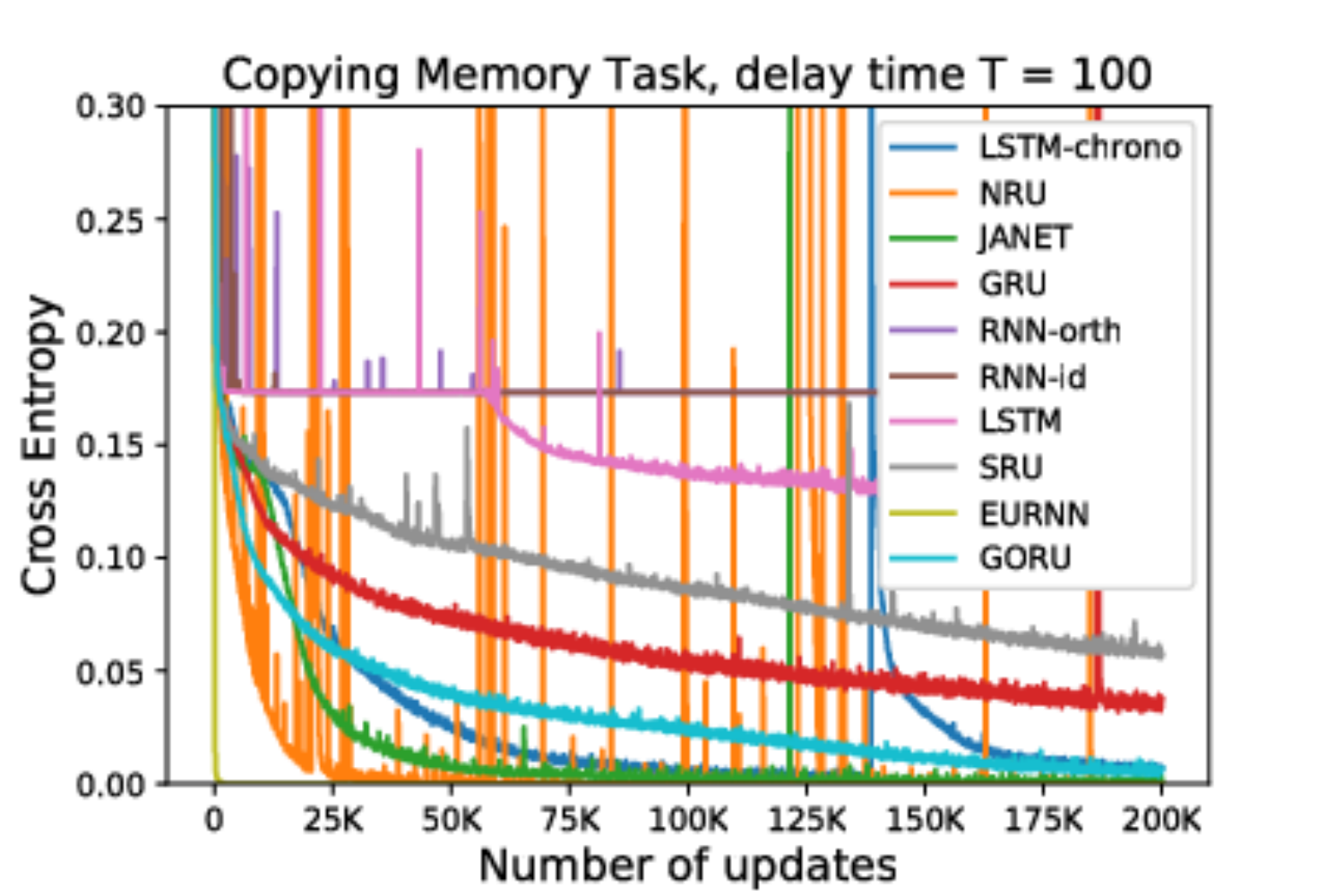} 
\end{minipage}% next column
\begin{minipage}{.45\textwidth}
\centering
\includegraphics[scale=0.55]{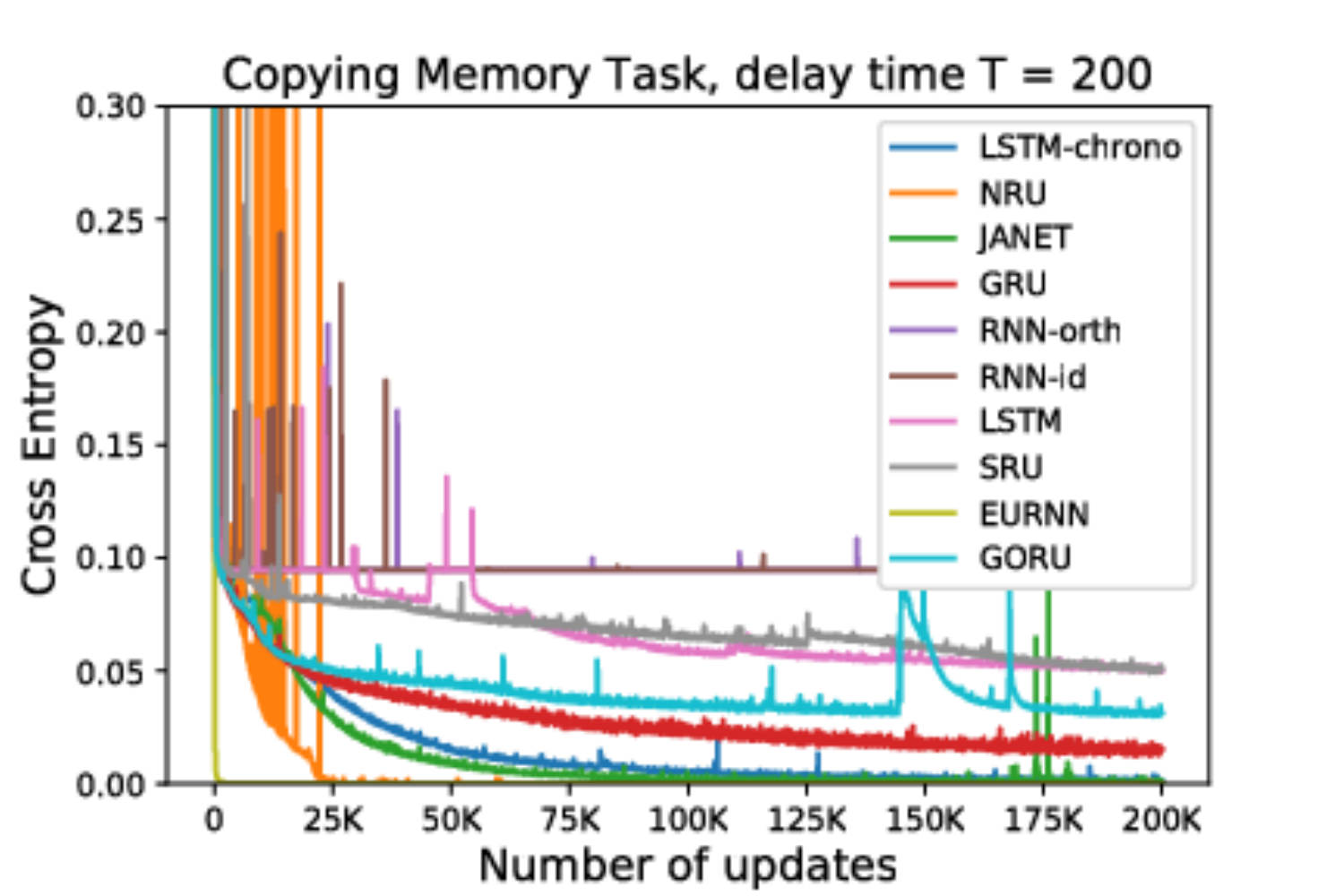}
\end{minipage}
\caption{Copying memory task for T = 100 (in left) and T = 200 (in right). Cross-entropy for random baseline : 0.17 and 0.09 for T=100 and T=200 respectively.}
\label{fig:copy}
\end{figure*}

The copying memory task was introduced in \cite{hochreiter1997} as a synthetic task to test the network's ability to remember information over many time steps. The task is defined as follows. Consider $n$ different symbols. In the first $k$ time steps, the network receives an initial sequence of $k$ random symbols from the set of $n$ symbols sampled with replacement. Then the network receives a ``blank" symbol for $T-1$ steps followed by a ``start recall" marker. The network should learn to predict a ``blank" symbol, followed by the initial sequence after the marker. Following \cite{arjovsky2016}, we set $n=8$ and $k=10$. The copying task is a pathologically difficult long-term dependency task where the output at the end of the sequence depends on the beginning of the sequence. We can vary the dependency length by varying the length of the sequence $T$. A memoryless model is expected to achieve a sub-optimal solution which predicts a constant sequence after the marker, for any input (referred to as the ``baseline" performance). The cross entropy for such a model would be $\frac{k\text{log}n}{T+2n}$ \cite{arjovsky2016}.

We trained all models with approximately the same number of parameters ($\sim$23.5k), in an online fashion where every mini-batch is dynamically generated. We consider $T=100$ and $T=200$. In Figure \ref{fig:copy} we plot the cross entropy loss for all the models. Unsurprisingly, EURNN solves the task in a few hundred updates as it is close to an optimal architecture tuned for this task \cite{henaff2016}. Conversely, RNN-orth and RNN-id get stuck at baseline performance. NRU converges faster than all other models, followed by JANET which requires two to three times more updates to solve the task.

We observed the change in the memory vector across the time steps (in the appendix). We can see that the network has learnt to add information into the memory in the beginning of the sequence and then it does not access the memory until it sees the marker. Then it makes changes in the memory in the last 10 time steps to copy the sequence.

\begin{figure*}[h]
\centering
\begin{minipage}{.5\textwidth}
\centering
\includegraphics[scale=0.55]{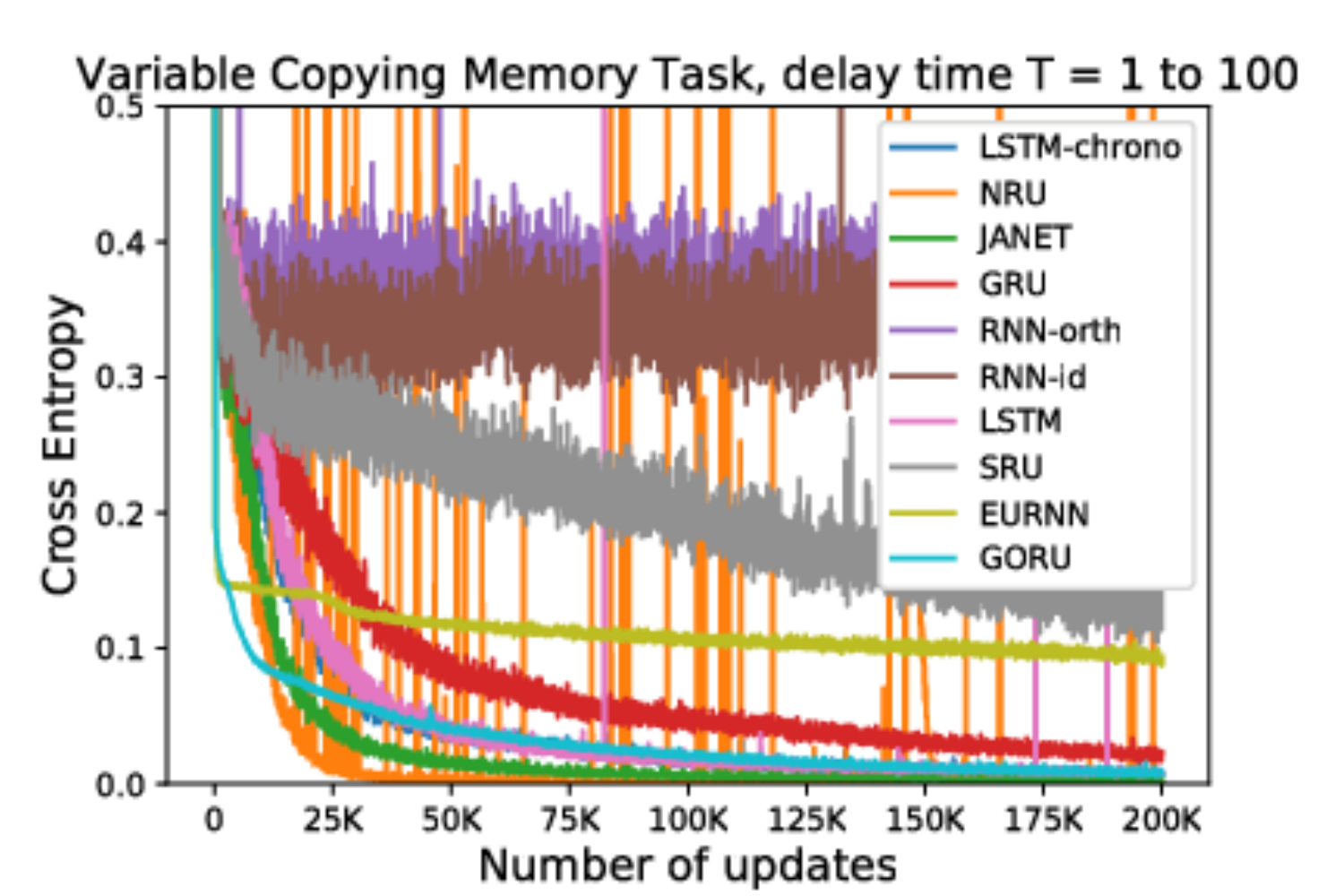} 
\end{minipage}% next column
\begin{minipage}{.45\textwidth}
\centering
\includegraphics[scale=0.55]{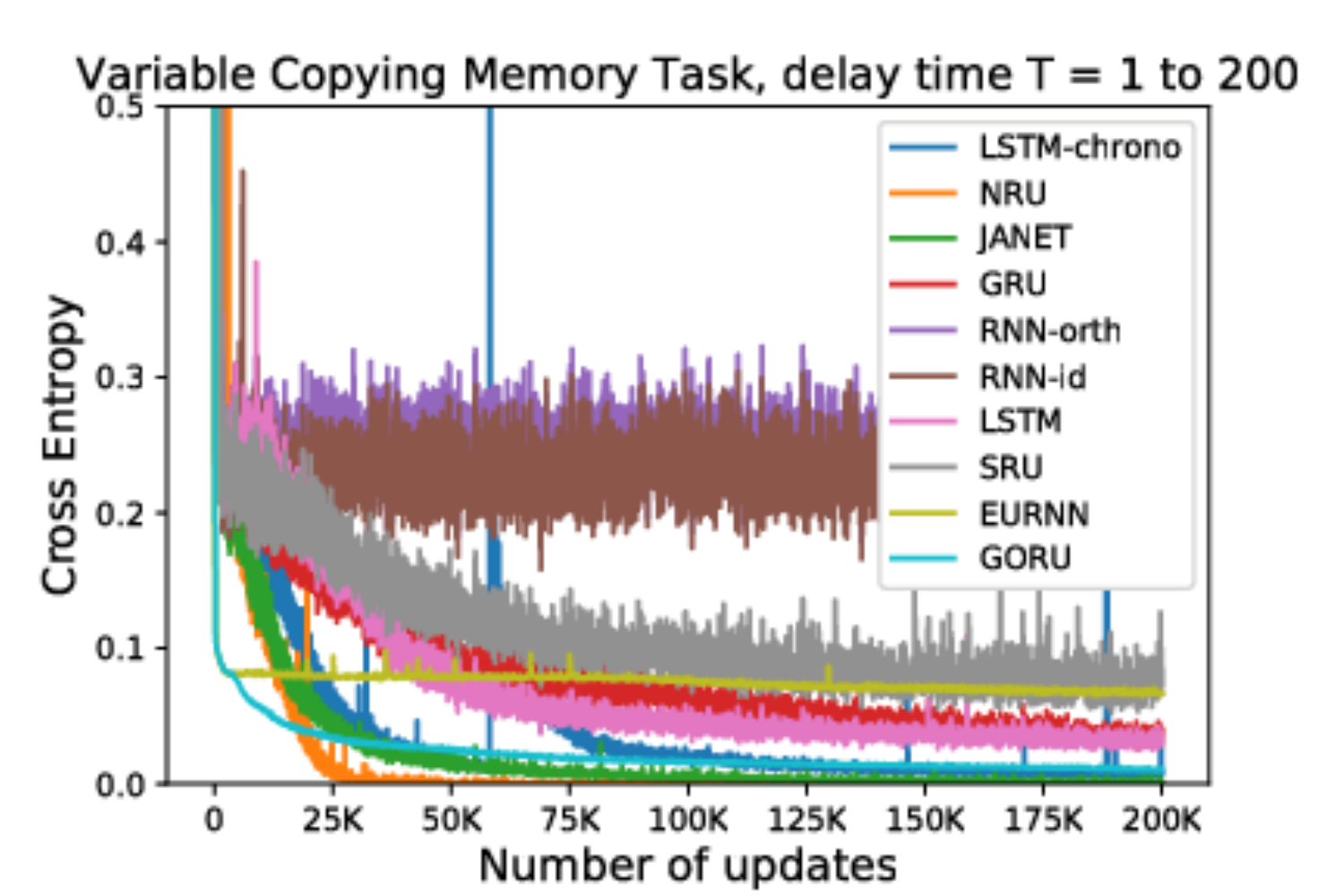}
\end{minipage}
\caption{Variable Copying memory task for T = 100 (in left) and T = 200 (in right).}
\label{fig:copy_var}
\end{figure*}

\begin{figure}[htbp]
    \centering
    \includegraphics[width=\linewidth]{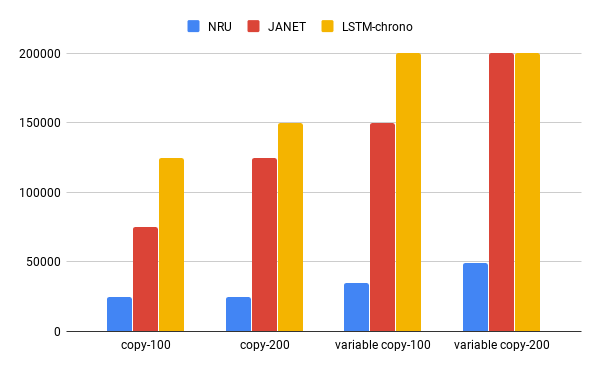}
    \caption{Comparison of top-3 models w.r.t the number of the steps to converge for different tasks. NRU converges significantly faster than JANET and LSTM-chrono.}
    \label{fig:model_comparison}
\end{figure}

We performed additional experiments following the observation by \cite{henaff2016} that gated models like the LSTM outperform a simple non-gated orthogonal network (similar to the EURNN) when the time lag $T$ is varied in the copying task. This variable length task highlights the non-generality of the solution learned by models like EURNN. Only models with a dynamic gating mechanism can solve this problem. In Figure-\ref{fig:copy_var}, we plot the cross-entropy loss for all the models. The proposed NRU model is the fastest to solve this task, followed by JANET, while the EURNN performs poorly, as expected according to \cite{henaff2016}. These two experiments highlight the fact that NRU can store information longer than other recurrent architectures. Unlike EURNN which behaves like a fixed clock mechanism, NRU learns a gating function to lock the information in the memory as long as needed. This is similar to the behaviour of other gated architectures. However, NRU beats other gated architectures mainly due to better gradient flow which results in faster learning. Figure-\ref{fig:model_comparison} shows that NRU converges significantly faster than its strongest competitors (JANET and LSTM-chrono) in all the 4 tasks.

\subsection{Denoising Task}

The denoising task was introduced in \cite{jing2017b} to test both the memorization and forgetting ability of RNN architectures. This is similar to the copying memory task. However, the data points are located randomly in a long noisy sequence. Consider again an alphabet of $n$ different symbols from which $k$ random symbols are sampled with replacement. These symbols are then separated by random lengths of strings composed of a ``noise" symbol, in a sequence of total length $T$. The network is tasked with predicting the $k$ symbols upon encountering a ``start recall" marker, after the length $T$ input sequence. Again we set $n=8$ and $k=10$. This task requires both an ability to learn long term dependencies and also an ability to filter out unwanted information (considered noise).

\begin{figure}[htbp]
\centering
\includegraphics[scale=0.55]{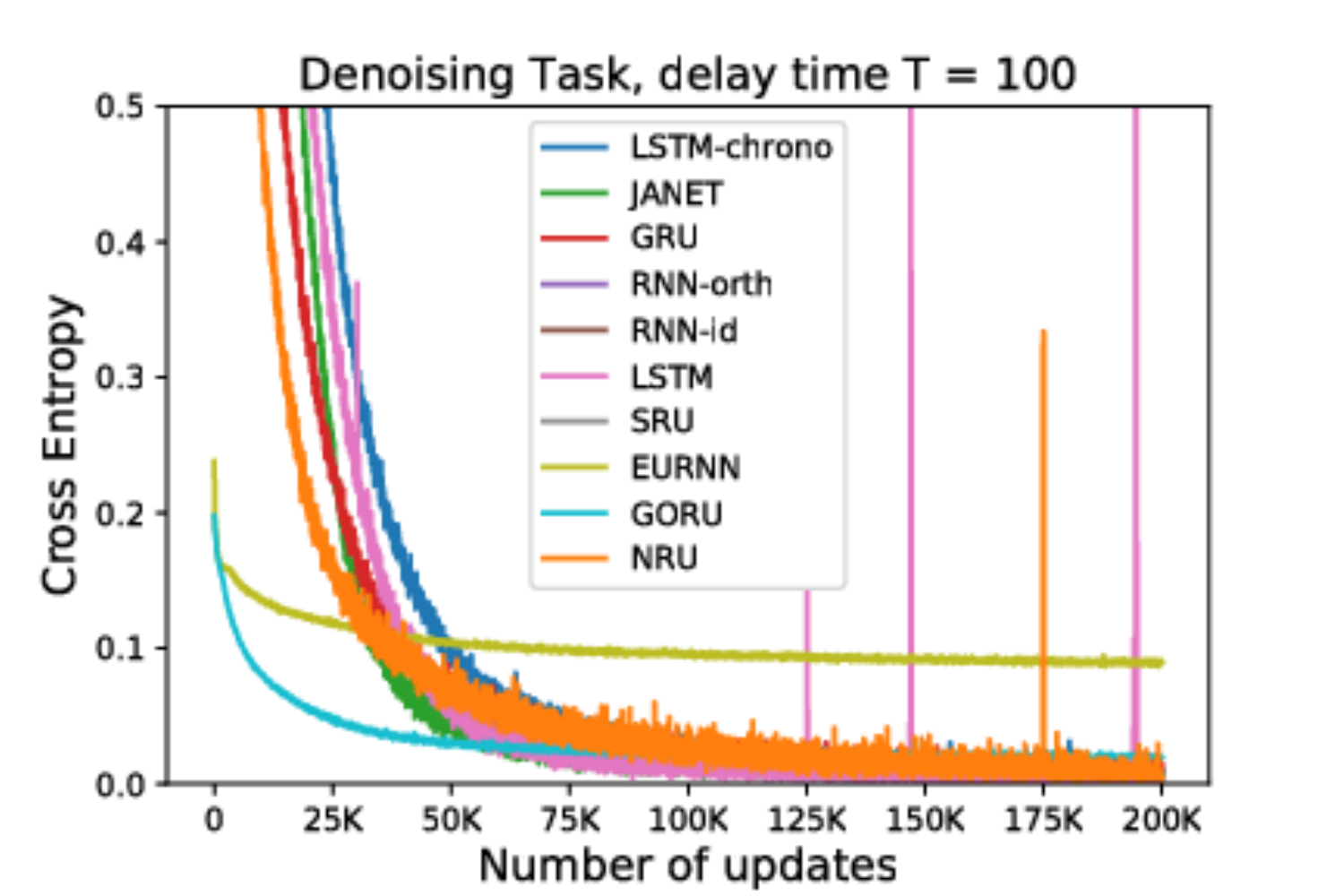}
\caption{Denoising task for T = 100.}
\label{fig:denoise}
\end{figure}

The experimental procedure is exactly the same as described for the copying memory task. In Figure \ref{fig:denoise} we plot the cross-entropy loss for all the models for $T=100$. In this task, all the models converge to the solution except EURNN. While NRU learns faster in the beginning, all the algorithms converge after approximately the same number of updates.

\begin{table}[htbp]
\centering
\caption{Bits Per Character (BPC) and Accuracy in test set for character level language modelling in PTB.}
\label{table:char}
\begin{tabular}{l|l|l}
\hline
\textbf{Model} & \textbf{BPC} & \textbf{Accuracy}\\
\hline
\hline
LSTM & 1.48 & 67.98  \\
LSTM-chrono & 1.51 & 67.41 \\
GRU & \textbf{1.45} & \textbf{69.07}\\
JANET & 1.48 & 68.50 \\
GORU & 1.53 & 67.60\\
EURNN & 1.77 & 63.00\\
NRU & 1.47 &  68.48\\
\hline
\end{tabular}
\end{table}

\subsection{Character Level Language Modelling}

Going beyond synthetic data, we consider character level language modelling with the Penn Treebank Corpus (PTB) \cite{marcus1993}. In this task, the network is fed one character per time step and the goal is to predict the next character. Again we made sure that all the networks have approximately the same capacity ($\sim$2.15M parameters). We use a batch size of 128 and perform truncated backpropagation through time, every 150 time steps. We evaluate according to bits per character (BPC) and accuracy.

All models were trained for 20 epochs and evaluated on the test set after selecting for each the model state which yields the lowest BPC on the validation set. The test set BPC and accuracy are reported in Table-\ref{table:char}. We did not add drop-out or batch normalization to any model. From the table, we can see that GRU is the best performing model, followed closely by NRU. 
We note that language modelling does not require very long term dependencies. This is supported by the fact that changing the additive memory updates in NRU to multiplicative updates does not hurt performance (it actually improves it). This is further supported by our observation that setting $T_{max}$ to 50 was better than setting $T_{max}$ to 150 in chrono-initialization for LSTM and JANET. All the best performing NRU models used ReLU activations in the writing and erasing heads.

\subsection{Permuted Sequential MNIST}

The Permuted Sequential MNIST task was introduced in \cite{le2015} as a benchmark task to measure the performance of RNNs in modelling complex long term dependencies. In a sequential MNIST task, all the 784 pixels of an MNIST digit are fed to the network one pixel at a time and the network must classify the digit in the 785th time step. Permuted sequential MNIST (psMNIST) makes the problem harder by applying a fixed permutation to the pixel sequence, thus introducing longer term dependencies between pixels in a more complex order.

\begin{figure}[htbp]
\centering
\includegraphics[scale=0.55]{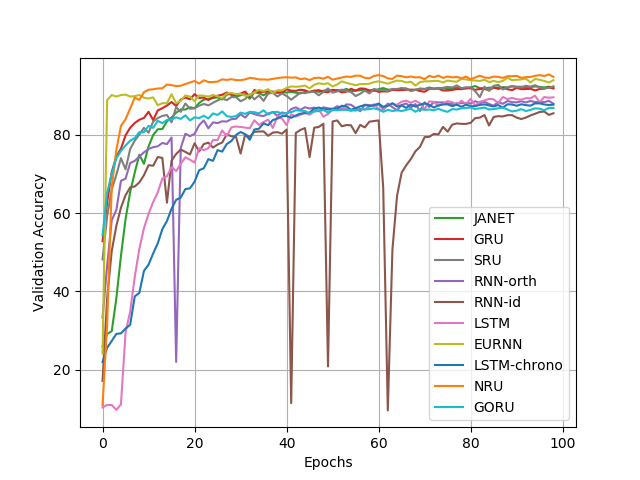}
\caption{Validation curve for psMNIST task.}
\label{fig:mnist}
\end{figure}

For this task, we used a batch size of 100. All the networks have approximately the same number of parameters ($\sim$165k). This corresponds to 200 to 400 hidden units for most of the architectures. Since the FFT-style algorithm used with EURNN requires few parameters, we used a large 1024 unit hidden state still achieving fewer parameters ($\sim$17k parameters) at maximum memory usage. We report validation and test accuracy in Table-\ref{table:mnist} and plot the validation curves in Figure-\ref{fig:mnist}. On this task, NRU performs better than all the other architectures, followed by the EURNN. The good performance of the EURNN could be attributed to its large hidden state, since it is trivial to store all 784 input values in 1024 hidden units. This model excels at preserving information, so performance is bounded by the classification performance of the output layer. While NRU remains stable, RNN-orth and RNN-id are not stable as seen from the learning curve. Surprisingly, LSTM-chrono is not performing better than regular LSTM.

\begin{table}[htbp]
\centering
\caption{Validation and test set accuracy for psMNIST task.}
\begin{tabular}{l|l|l}
\hline
\textbf{Model} & \textbf{valid} & \textbf{test}\\
\hline
\hline
RNN-orth & 88.70 & 89.26  \\
RNN-id & 85.98 & 86.13  \\
LSTM & 90.01 & 89.86  \\
LSTM-chrono & 88.10  & 88.43  \\
GRU & 92.16 & 92.39  \\
JANET & 92.50 & 91.94  \\
SRU & 92.79 & 92.49  \\
EURNN & 94.60 & 94.50  \\
GORU & 86.90 & 87.00  \\
NRU & \textbf{95.46} & \textbf{95.38}\\
\hline
\end{tabular}
\label{table:mnist}
\end{table}

\subsection{Model Analysis}
\textbf{Stability}: While we use gradient clipping to limit exploding gradients in NRU, we observed gradient spikes in some of our experiments. We suspect that the network recovers due to consistent gradient flow. To verify that the additive nature of the memory does not cause memory to explode for longer time steps, we performed a sanity check experiment where we trained the model on the copy task (with $T$=100, 200, and 500) with random labels and observed that training did not diverge. We observed some instabilities with all the models when training with longer time steps ($T$ = 2000). With NRU, we observed that the model converged faster and was more stable with a higher memory size. For instance, our model converged almost twice as early when we increased the memory size from 64 to 144.

\noindent\textbf{Forgetting Ability}:
We performed another sanity check experiment to gauge the forgetting ability of our model. We trained it on the copy task but reset the memory only once every $k$= 2, 5, and 10 examples and observed that NRU learned to reset the memory in the beginning of every example.

\noindent\textbf{Gradient Flow}:
% TO DO : add plots.
To give an overview of the gradient flow across different time steps during training, we compared the total gradient norms in the copying memory task for the top 3 models: NRU, JANET and LSTM-Chrono (in the appendix). We see that the NRU’s gradient norm is considerably higher during the initial stages of the training while the other model’s gradient norms rise after about 25k steps. After 25k steps, the drop in the NRU gradient norm coincides with the model’s convergence, as expected. We expect that this ease of gradient flow in NRU serves as an additional evidence that NRU can model long-term dependencies better than other architectures.

\section{Conclusion}
In this paper, we introduce Non-saturating Recurrent Units (NRUs) for modelling long term dependencies in sequential problems. The gating mechanism in NRU is additive (like in LSTM and GRU) and non-saturating (unlike in LSTM and GRU). This results in better gradient flow for longer durations. We present empirical evidence in support of non-saturating gates in the NRU with (1) improved performance on long term dependency tasks, (2) higher gradient norms, and (3) faster convergence when compared to baseline models. NRU was the best performing general purpose model in all of the long-term dependency tasks that we considered and is competitive with other gated architectures in short-term dependency tasks. This work opens the door to other potential non-saturating gated recurrent network architectures.

We would also like to apply NRU in several real world sequential problems in natural language processing and reinforcement learning where the LSTM is the default choice for recurrent computation.

%\small
\bibliography{refs}
\bibliographystyle{aaai}

\onecolumn
\section{Appendix}

\subsection{Memory Visualization}
\label{appendix_mem_viz}
\begin{figure*}[htbp]
\centering
\includegraphics[scale=0.33]{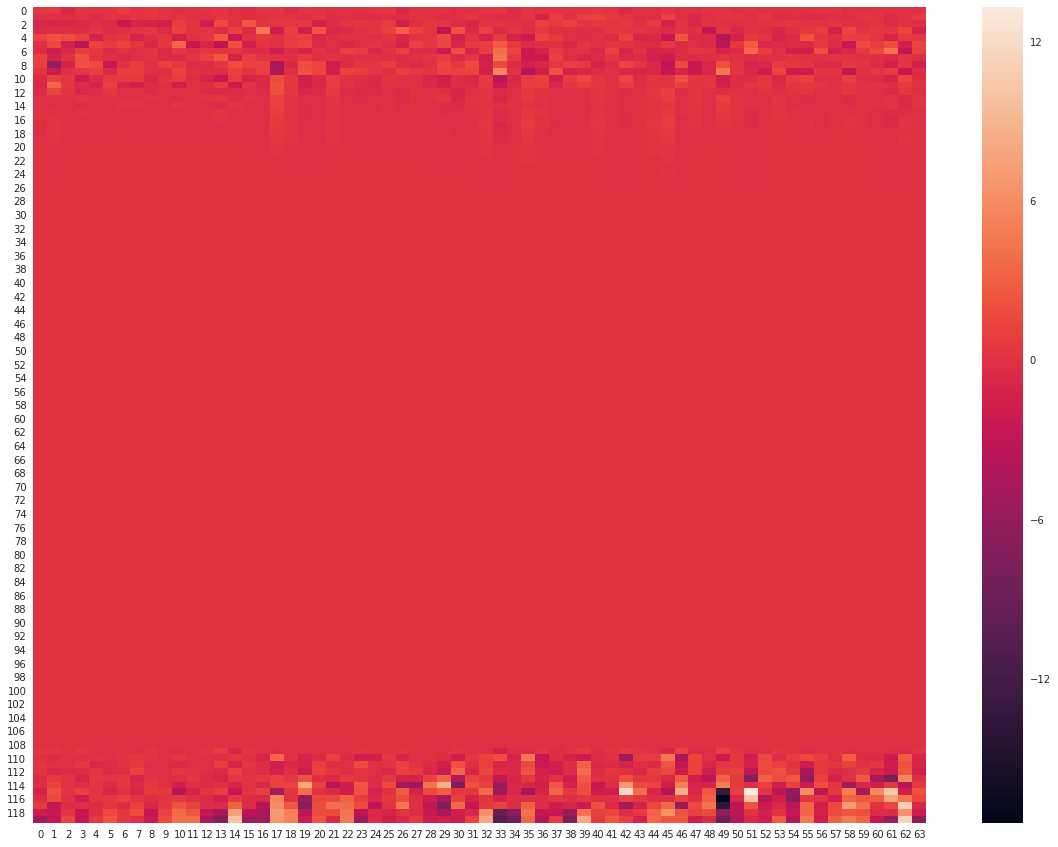}
\caption{Change in the content of the NRU memory vector for the copying memory task with T=100. We see that the network has learnt to use the memory in the first 10 time steps to store the sequence. Then it does not access the memory until it sees the marker. Then it starts accessing the memory to generate the sequence.}
\label{fig:mem_viz2}
\end{figure*}

In Figure \ref{fig:mem_viz2}, we visualize the change in the content of the NRU memory vector for the copying memory task with T=100. We see that the network has learnt to use the memory in the first 10 time steps to store the sequence. Then it does not access the memory until it sees the marker. Then it starts accessing the memory to generate the sequence.

\subsection{Gradient Norm Comparison}

\begin{figure}[htbp]
\centering
\includegraphics[scale=0.3]{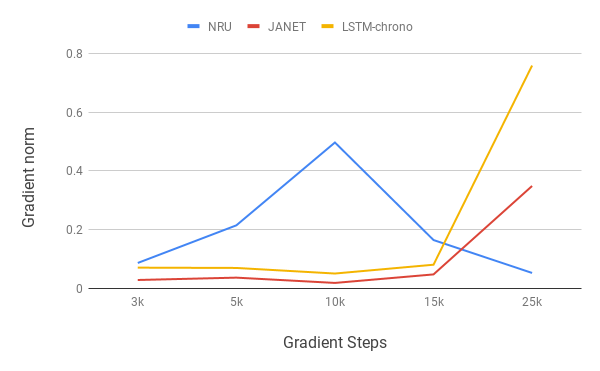}
\caption{Gradient norm comparison with JANET and LSTM-chrono across the training steps. We observe significantly higher gradient norms for NRU during the initial stages compared to JANET or LSTM-chrono. As expected, NRU's gradient norms decline after about 25k steps since the model has converged.}
\label{fig:gradient_norm_vs_steps}
\end{figure}

\subsection{Hyper-parameter Sensitivity Analysis}

\begin{figure*}[htbp]
\minipage{0.32\textwidth}
  \includegraphics[width=\linewidth]{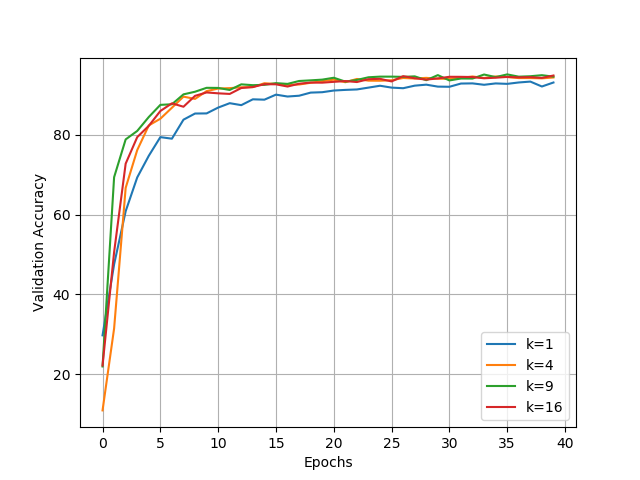}
\endminipage\hfill
\minipage{0.32\textwidth}
  \includegraphics[width=\linewidth]{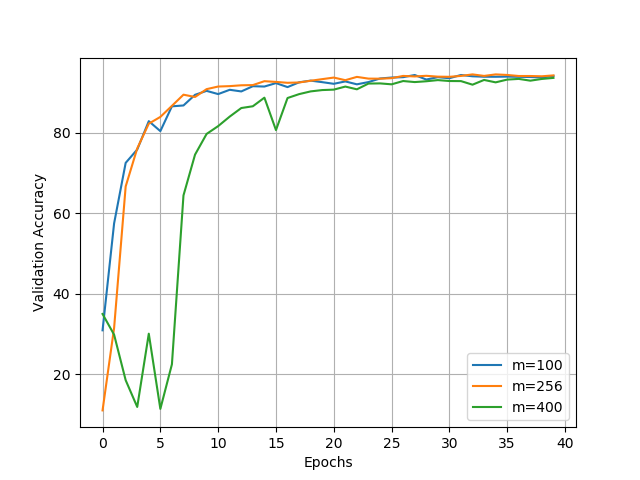}
\endminipage\hfill
\minipage{0.32\textwidth}%
  \includegraphics[width=\linewidth]{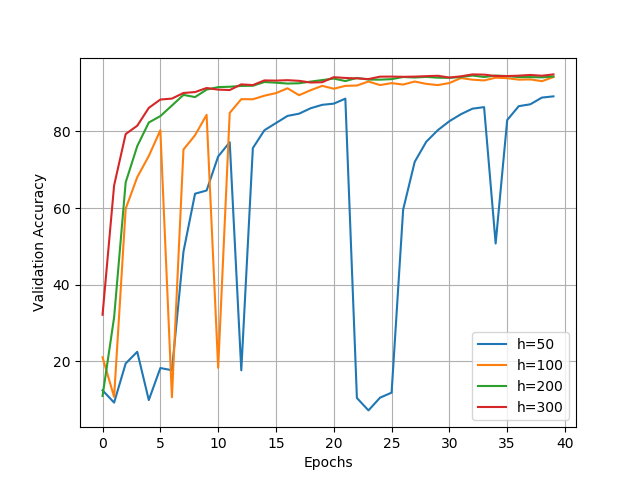}
\endminipage
 \caption{Effect of varying the number of heads (left), memory size (middle), and hidden state size (right) in psMNIST task.}
 \label{hyper}
\end{figure*}

NRU has three major hyper-parameters: memory size ($m$), number of heads ($k$) and hidden state size ($h$). To understand the effect of these design choices, we varied each hyper-paramteter by fixing other hyper-parameters for the psMNIST task. The baseline model was $m=256, k=4, h=200$. We varied $k \in \{1, 4, 9 , 16\}$, $m \in \{100, 256, 400\}$, and $h \in \{50, 100, 200, 300\}$. We trained all the models for a fixed number of 40 epochs. Validation curves are plotted in Figure-\ref{hyper}. It appears that increasing $k$ is helpful, but yields diminishing returns above 4 or 9. For a hidden state size of 200, a larger memory size (like 400) makes the learning difficult in the beginning, though all the models converge to a similar solution. This may be due to the increased number of parameters. On the other hand, for a memory size of 256, small hidden states appear detrimental.

\end{document}